\documentclass[letterpaper, 10 pt, conference]{ieeeconf}  

\IEEEoverridecommandlockouts                     
\def\IEEEIROS{{\rm I\kern-.1em{\sc eee}\kern-.05em{\sc iros}}}

\usepackage[dvipsnames]{xcolor} 
\usepackage{graphicx} 
\usepackage{array} 
\usepackage{multirow}
\usepackage{array} 
\usepackage{amsmath} 
\usepackage{amsfonts} 
\usepackage{cite}
\usepackage{siunitx}
\usepackage{xcolor}
\usepackage{soul}
\usepackage{hyperref}
\hypersetup{
    colorlinks=true,
    linkcolor=black,     
    citecolor=black,     
    filecolor=black,     
    urlcolor=blue       
}

\definecolor{mycolor}{rgb}{0.1,0.2,0.5}
\definecolor{brightblue}{rgb}{0.0, 0.5, 1.0} 
\sethlcolor{yellow} 

\title{\LARGE \bf
Motor Imagery Teleoperation of a Mobile Robot Using a Low-Cost Brain-Computer Interface for Multi-Day Validation}

\author{Yujin An$^{*}$ and Daniel Mitchell$^{*}$, John Lathrop, David Flynn, and Soon-Jo Chung
\thanks{*Equal contribution. Y. An, J. Lathrop, and S.J. Chung are with the Division of Engineering and Applied Science, California Institute of Technology. {yan2, jlathrop, sjchung(@caltech.edu)}. D. Mitchell and D. Flynn are with the Autonomous Systems and Connectivity Division, University of Glasgow, Glasgow, U.K. {Daniel.Mitchell, David.Flynn(@Glasgow.ac.uk)}. This project was in part funded by the Amazon AWS and Carver Mead Discovery Funds (Caltech), and the MacRobertson, Mobility scholarship and the Scottish Int. Edu. Trust (U. Glasgow).}}%

\begin{document}

\maketitle
\thispagestyle{empty}
\pagestyle{empty}

\begin{abstract}
 Brain-computer interfaces (BCI) have the potential to provide transformative control in prosthetics, assistive technologies (wheelchairs), robotics, and human-computer interfaces. While Motor Imagery (MI) offers an intuitive approach to BCI control, its practical implementation is often limited by the requirement for expensive devices, extensive training data, and complex algorithms, leading to user fatigue and reduced accessibility. In this paper, we demonstrate that effective MI-BCI control of a mobile robot in real-world settings can be achieved using a fine-tuned Deep Neural Network (DNN) with a sliding window, eliminating the need for complex feature extractions for real-time robot control. The fine-tuning process optimizes the convolutional and attention layers of the DNN to adapt to each user's daily MI data streams, reducing training data by 70\% and minimizing user fatigue from extended data collection. Using a low-cost ($\sim$\$3k), 16-channel, non-invasive, open-source electroencephalogram (EEG) device, four users teleoperated a quadruped robot over three days. The system achieved 78\% accuracy on a single-day validation dataset and maintained a 75\% validation accuracy over three days without extensive retraining from day-to-day. For real-world robot command classification, we achieved an average of 62\% accuracy. By providing empirical evidence that MI-BCI systems can maintain performance over multiple days with reduced training data to DNN and a low-cost EEG device, our work enhances the practicality and accessibility of BCI technology. This advancement makes BCI applications more feasible for real-world scenarios, particularly in controlling robotic systems.

\end{abstract}
\section{INTRODUCTION}



Brain Computer Interfaces (BCI) for robotic applications are an emerging technology gaining significant traction, unlocking new levels of control and enhancing human abilities \cite{SCIENCE_ROBOTICS_BCI, SCI_ROBOT_EEG_EOG}. Synchronizing the human brain with external devices, such as wheelchairs, mobile robots and robot arms, holds the potential to drastically reduce barriers for paralyzed individuals, enabling them to teleoperate their robots via their thoughts \cite{7989777_Real_Time_ICRA,10423905, Mitchell_SARESE, Mitchell_SSOSA}. BCIs collect brain responses, process the signals, and convert them into commands for robots or end effectors to perform the user's intended actions \cite{10144535,8368080, Silversmith2021}.

Among the various brain machine interface options for teleoperating robots, non-invasive electroencephalogram (EEG) is particularly promising due to its high temporal resolution, user-friendly operation, and safety features. Most EEG recording devices are expensive, equipped with numerous electrodes, and uncomfortable due to their wired configurations. More recently, low-cost and portable EEG headsets have the potential to improve accessibility for individuals where access to accurate BCI would positively impact their daily lives, making the technology more practical and widely available \cite{9606552}.

A specific EEG-based BCI approach known as Motor Imagery (MI) has gained attention for allowing users to control external devices by mentally simulating physical body movements, such as those of the hands and feet. This method has a higher acceptance from end-users because it is more intuitive and causes less fatigue compared to methods that rely on external stimulation \cite{Tonin_BCI}. MI has been applied to real-world robotic manipulators and mobile robot control scenarios \cite{Simanto_2021, Baniqued2021}. 

Despite the growing attention in Deep Neural Networks (DNN) for MI decoding—due to their reduced need for extensive pre-processing \cite{AMIN2019542}—practical implementation remains challenging, limited by extensive training data, complex processes, and expensive devices, leading to user fatigue and reduced accessibility. Firstly, extensive training data causes user fatigue and potential degradation of model performance post-validation. Secondly, inter-individual differences and day-to-day variability in brain patterns hinder high-accuracy decoding and complicate model generalizability \cite{chau2024generalizability, doi:10.1126/scirobotics.abd1911}. Thirdly, systems needing higher accuracy often require more EEG sensors, increasing costs and complexity\cite{app10217453}. These challenges are particularly pronounced in real-time mobile robot control, where high-dimensional commands pose significant difficulties\cite{doi:10.1126/scirobotics.abd1911}. Consequently, many MI-controlled robot studies use few EEG commands—such as single-direction linear velocity tracking or simple left-right turning—that are validated within a single day, highlighting the difficulty in effectively decoding more complex MI signals\cite{XU20233, 7109829}. Collectively, these factors limit daily use and hinder the accelerated development of MI-BCIs. 

\begin{figure*}[!t]
\vspace*{0.3cm}
    \centering
    \includegraphics[width=0.95\linewidth]{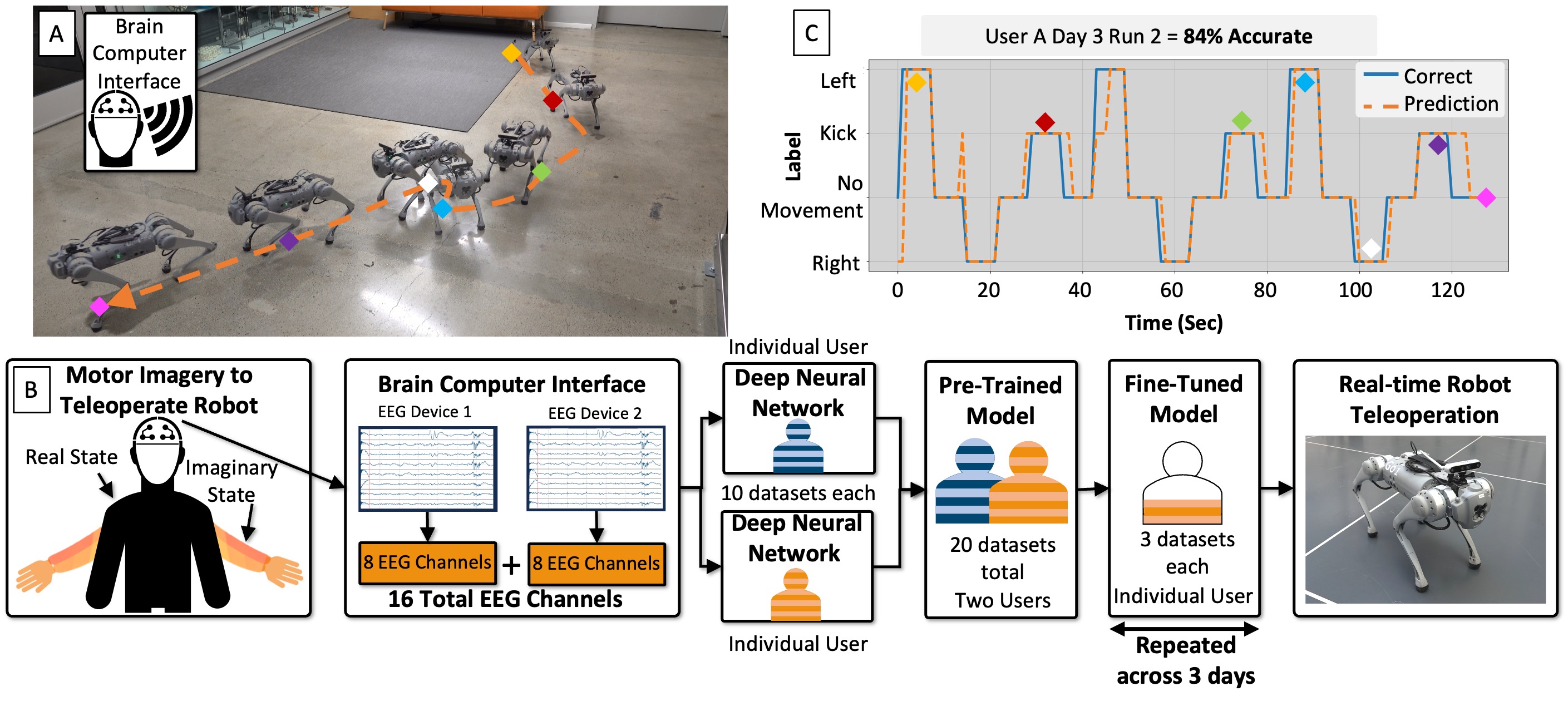}
    \vspace*{-0.5cm}
    \caption{The proposed approach for teleoperating the mobile robot via the BCI. (A) The robot during a run being teleoperated via the BCI device. (B) An overview of the proposed methodology utilizing a 16-channel BCI via MI with DNN and fine-tuning. Users A and B, and Users C and D were paired to create pre-trained models. Each user's fine-tuned model was developed by further training the pre-trained model that included their data. (C) The visualization of the real-time EEG decoding output where the real robot can be identified in (A) with corresponding colored diamonds matching each segment. A video of this research can be accessed: \href{https://youtu.be/1rR7YFRAJhs?si=6Yj7lB1G7XL-0VAh}{https://youtu.be/1rR7YFRAJhs?si=6Yj7lB1G7XL-0VAh}.}
    \label{fig_Novelty}
    \vspace*{-0.5cm}
\end{figure*}

Here, we present a low-cost EEG-BCI system that employs a fine-tuned DNN with a sliding window, eliminating the need for complex feature extraction methods and maintaining high accuracy when conducting 4-class MI tasks for a mobile robot. The fine-tuning process optimizes the model for each user and day by retraining only the initial convolutional and final linear layers of a pre-trained model. This allows the system to adapt to the unique EEG patterns of individual users and handle day-to-day variations in neural signals. The system was validated with two pairs of users (four total), as shown in Fig. 1. The main contributions include: (A) reducing dataset size by 70\% while maintaining 75\% average accuracy for the validation over three days; (B) achieving 62\% accuracy for real-time robot control with short calibration; and (C) using a low-cost, open-source 16-channel EEG system, significantly lowering hardware and software costs. These findings offer insights into designing practical MI-BCI systems that efficiently manage day-to-day user variability.

\section{RELATED WORK}

Non-invasive BCIs measure brain activity without surgery, offering greater safety and lower costs. The main types are P300, Steady-State Visually Evoked Potentials (SSVEPs), and Motor Imagery (MI). While P300 and SSVEP have their advantages, we focus on MI, which enables asynchronous control of manipulators and mobile robots without additional hardware. Although many BCI studies demonstrate more than four commands, they often employ hybrid approaches combining different techniques \cite{ZHANG202112, s21186285, zhang2023NOIR_MIT}. This review specifically examines MI techniques, focusing on key aspects: the number of days validated, the duration of data collection to build accurate models, the number of MI commands achieved, and model performance across multiple sessions or days for sustained high accuracy. Table \ref{table_Lit_Review} presents a comparative analysis of real-world deployments of MI-BCI for robot manipulators and mobile robots, including key results from this article for comparison. The remainder of this section summarizes key contributions from the state of the art.

\begin{table*}[!t]
\vspace*{+0.3cm}
\caption{Comparative Analysis of Non-invasive  MI-BCI for Real-time Control of Real-world Manipulators and Mobile Robots}
\label{table_Lit_Review}
\centering
\begin{tabular}{|>{\centering\arraybackslash}p{3.4cm}|>{\centering\arraybackslash}p{1.1cm}|>{\centering\arraybackslash}p{1.4cm}|>{\centering\arraybackslash}p{0.6cm}|>{\centering\arraybackslash}p{1.4cm}|>{\centering\arraybackslash}p{1.4cm}|>{\centering\arraybackslash}p{1.6cm}|>{\centering\arraybackslash}p{1.3cm}|>{\centering\arraybackslash}p{1.4cm}|}
\hline

\multirow{2}{*}{\textbf{Article, Year, Ref}} & {\textbf{Validated across No. Days}} & \textbf{Decoding Model} & \textbf{No. Users} & \textbf{No. EEG Nodes} & \textbf{No. Commands} & \textbf{User Data Collection Duration (hours)} & \multicolumn{2}{c|}{\textbf{Accuracy (\%)}} \\
\cline{8-9} 
 &  &  &  &  &  &  & \multicolumn{1}{c|}{\textbf{Validation}} & \textbf{Real-world Robot} \\ \hline
 
Teleop. Robot Ctrl. Tactile Feedback, 2020, \cite{electronics9010174} & 1 & Linear Regression & 6 & 9 & 4 & 3.8 & Unavailable$^{\mathrm{a}}$ & Unavailable$^{\mathrm{a}}$ \\ \hline

Assistive Shared Ctrl.; Robot Arm, 2021, \cite{9636261} & 1 & Gaussian Classifier & 1 & 8 & 4 & Unavailable & {Unavailable$^{\mathrm{a}}$} & Unavailable$^{\mathrm{a}}$ \\ \hline

Optimizing MI Param. Robot Arm Ctrl., 2022, \cite{brainsci12070833} & 1 & LDA & 7 & 64 & 3 & 0.4 & \multicolumn{1}{c|}{Unavailable$^{\mathrm{b}}$} & Unavailable$^{\mathrm{a}}$ \\ \hline

Brain-Actuated Humanoid Robots: Asynch. Direct Ctrl Using EEG-Based BCI, 2012, \cite{6214617} & 1 & LDA \& QDA & 5 & 21 & 4 & 1 & 80 & Unavailable$^{\mathrm{a}}$ \\ \hline

A BCI Telepresence Robot for People With Severe Motor Disabilities, 2015, \cite{7109829} & 1$^{\mathrm{c}}$ & Gaussian Classifier & 19 & 16 & 2 & 6$^{\mathrm{d}}$ & Unavailable$^{\mathrm{a}}$ & Unavailable$^{\mathrm{a}}$ \\ \hline

Cont. shared control of a mobile robot with BCI, 2023, \cite{XU20233} & 1 & Linear Regression & 6 & 2 & 2 & 4.83 & Unavailable$^{\mathrm{e}}$ & Unavailable$^{\mathrm{a}}$ \\ \hline

MI BCI and LiDAR-Based Wheelchair, 2023, \cite{Wheelchair_BCI_10144535} & 1 & DNN & 5 & 20 & 3 & 0.9 & 83$^{\mathrm{f}}$ & Unavailable$^{\mathrm{a}}$ \\
 \hline
 
\textbf{Our Approach: MI Teleop. of a Mobile Robot Using a Low-Cost BCI for Multi-Day Valid., 2024} \vspace{-7mm} & 4 & DNN & 4 & 16 & 4 & 2.16 & 78 & 57  \\ \cline{7-9} 
 &  &  &  &  &  & 0.65 & 75 & 62 \vspace{4mm}\\ \hline

\multicolumn{9}{|p{17.0cm}|}{\raggedright 
$^{\mathrm{a}}$No indication of MI accuracy and they used a different method to validate the accuracy of the system.\\
$^{\mathrm{b}}$User 1 and 3 demonstrated 70\% accuracy, however, the author indicated poor results from other users but never indicated the overall accuracy.\\
$^{\mathrm{c}}$User 1 and 4 went on holiday, necessitating repeat online training on a different day.\\
$^{\mathrm{d}}$A minimum of two sessions were assumed as the author did not specify, only detailing 3 hours once or twice a week.\\
$^{\mathrm{e}}$Achieved 78\% on cursor control tasks before training, improving to 97\% after training.\\
$^{\mathrm{f}}$Users could not proceed to real-world testing until 75\% had been achieved in validation.
}

\\ \hline 
\end{tabular}
\vspace*{-0.6cm}
\end{table*}

MI-BCI teleoperation of manipulator arms has been demonstrated in several studies. In \cite{electronics9010174}, a small desktop manipulator with tactile feedback was controlled using linear regression; users required an average of 3.8 hours of training to achieve a 66\% task success rate in task completion when moving the robotic arm to the correct location, although MI command classification accuracy was not reported. A shared control architecture for a single user was validated in \cite{9636261}, where EEG signals were decoded using a Gaussian classifier, resulting 4 out of 10 errors; the training duration was not specified. Another study \cite{brainsci12070833} used 64 EEG electrodes to teleoperate a manipulator with three commands; four sets of training data over 0.4 hours were required, involving feature extraction and Linear Discriminant Analysis (LDA), necessitating high complex process. 

Next, controlling mobile robots via MI-BCI presents significant challenges. In \cite{6214617}, four distinct commands were classified by a human operator to control a NAO humanoid robot using LDA and Quadratic Discriminant Analysis (QDA). Users were asked for substantial training before real-world testing; users had to achieve a minimum offline accuracy of 75\% before progressing to real-time control, requiring additional training sessions if this threshold was not met. The study reported average offline and online accuracies of 78.6\% and 80\%, respectively, but these do not reflect the MI command accuracy during actual robot control. Moreover, experiments were conducted in a controlled environment with the robot operating at a very slow pace, limiting generalizability to more dynamic settings. Similarly, \cite{7109829} deployed a BCI-controlled mobile robot for end-users with severe paralysis, utilizing the two highest-performing MI commands per user with complex feature extraction (e.g., feet and right-hand MI versus feet and left-hand MI). Users guided the robot to various target goals in a simple environment via a Gaussian classifier. However, this approach required long training durations of at least 3 hours to achieve sufficient control for a small number of commands and involved complex pre-processing to select EEG channels for users.

The classification of MI for two commands—both hand movement and relaxation—was presented in \cite{XU20233}. In the training phase, users controlled a cursor, and for real-world testing, the approach employed a shared control strategy with a linear regression model to avoid obstacles in a corridor using two command inputs. Similarly, \cite{Wheelchair_BCI_10144535}
 introduced a brain-controlled wheelchair utilizing shared autonomy and a Deep Neural Network (DNN) for three classes of MI commands. Users could navigate forward, right, and left, but stopping the wheelchair depended on their concentration state.
 
MI-BCI faces key challenges, including the lack of validation for the reliability of BCI-controlled robotic systems over multiple days, which is essential for consistent daily use and is the focus of this research.  In addition, the current methods often require extensive training datasets for the user to proficiently utilize their brain signals. Also, there are trade-offs between the number of required EEG nodes, the number of commands, and system accuracy. Increasing EEG nodes typically raises costs and system complexity. While recent studies have focused on classifying 2–4 MI-BCI commands, significant challenges remain in decoding the large number of commands with high accuracy and simpler processing. To date, researchers have typically used other methods, such as SSVEP or P300, instead of MI for real-time robot control to overcome this issue \cite{ZHANG202112, s21186285}. 

Furthermore, a standardized methodology is essential to validate MI-BCI commands for mobile robots by quantifying accuracy during both validation and real-world deployment. Current studies typically use metrics such as task success rate, duration, and completion time \cite{electronics9010174, XU20233,Wheelchair_BCI_10144535}. This article proposes a standardized method for future MI-BCI research to measure robot control accuracy during teleoperation and classify four commands while reducing training time through a DNN with a sliding window for real-time control, as shown in Table \ref{table_Lit_Review}.

\section{METHODOLOGY}

This section details the procedure for the EEG-BCI device. Users were informed of the study's general features in advance, but no extensive user practice was provided. The investigation aimed to validate a fine-tuned model to reduce the need for large datasets when calibrating for real-world teleoperation of a mobile robot.

\subsection{BCI Specification}

Two Unicorn Naked BCI devices (g.tec, Austria), each with 8 channels, connected to a laptop via Bluetooth 2.1 to create a low-cost and wireless EEG headset. The 16 total EEG node positions can be viewed in Fig. \ref{fig_Trio}A, where the 10-20 international system was followed. The first device utilized C3, Cz, C4, AF3, AF4, P3, P4, POz and second device positioned nodes on C5, FC3, CP3, C1, C2, FC4, CP4, C6. Ear nodes were connected together, bone nodes were also connected together and a ground belt was also connected to the wrists of the users. When calibrating the BCI device, each EEG node was kept between $ \pm $50 \si{\micro\volt} with the recommended EEG gel. The EEG signals were sampled with a 24-bit resolution and 250Hz per channel. Each user was situated in a control room (Fig. \ref{fig_Trio}B) during the investigation to ensure they were free from distractions or disturbances. A real-world robot was used to ensure accurate model creation and application during data collection and testing. An overview of the main electronics for the BCI device can be viewed in Fig. \ref{fig:Procedure_Merged}A. 

\subsection{Data Collection}
The user imagined each motor imagery task for 7 seconds without performing any actual movement:
\begin{itemize}
    \item Movement of the Right hand (\textbf{R}),
    \item Movement of the Left hand (\textbf{L}),
    \item Kicking (\textbf{K}),
    \item No Movement (\textbf{N})
\end{itemize}

One dataset (13 minutes of tasks) consists of six runs. A single run comprises of nine sequences of motor imagery tasks, denoted as $c_i \in \{\text{NR}, \text{NL}, \text{NK}\}$, where each task within a run is presented in a random order to mitigate any sequential learning effects. Each task was marked on the screen with a corresponding small arrow, while the no movement state was indicated by a standard crosshair indicating a user to relax their visualized arm, resulting in a stationary robot. To ensure participants maintained concentration and to mitigate fatigue, a 30-second rest period was provided between consecutive runs. Additional rest periods were enforced between different datasets to allow for full recovery and sustain the accuracy of MI representation throughout the investigation.


\subsection{Study Procedure}

\begin{figure}[!t]
    \centering
    \vspace*{0.3cm}
    \includegraphics[width=0.7\linewidth]{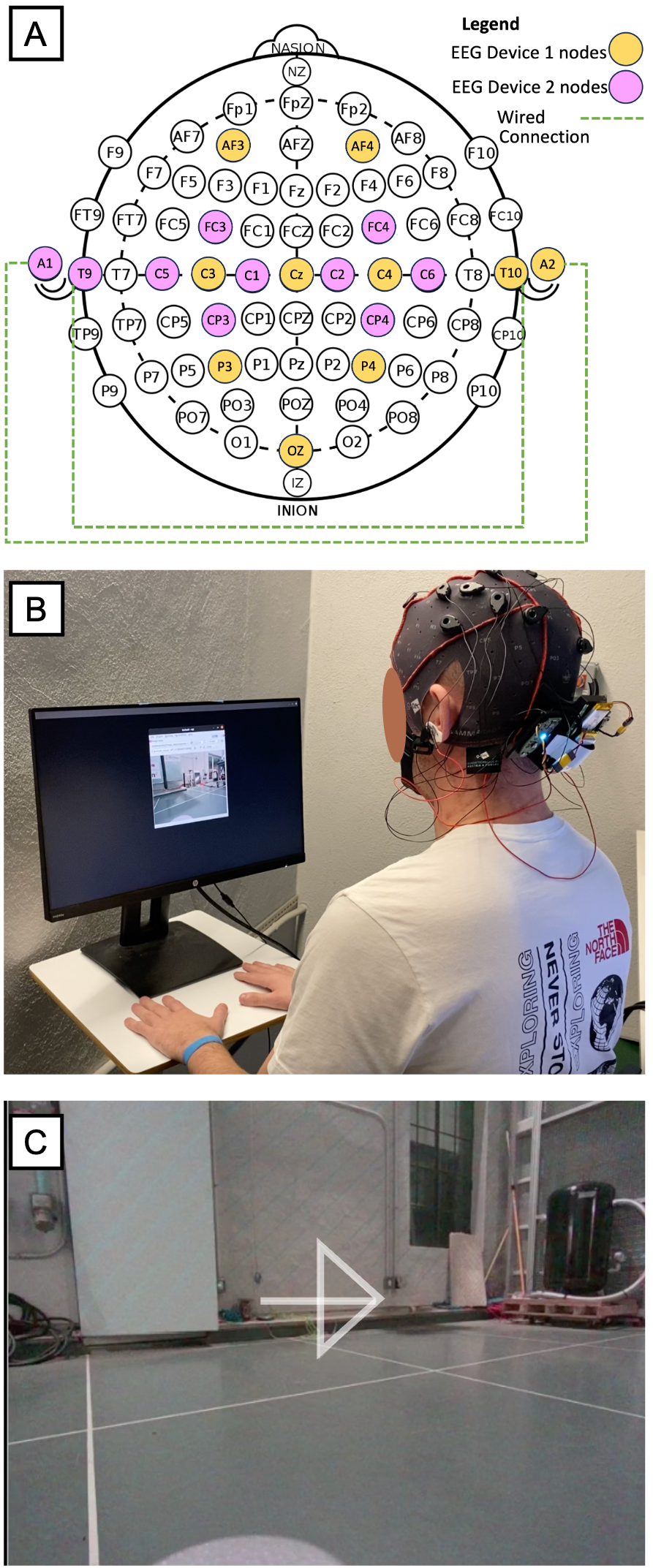}
    \caption{(A) EEG node positions used during the investigation for the 16 channel device. (B) A view of the BCI, GUI and user during the investigation. (C) Point of View from the camera mounted on quadruped robot displayed to the user from the GUI.  The crosshair size has been increased slightly compared to the real implementation for improved visualization.}
    \label{fig_Trio}
    \vspace*{-0.2cm}
\end{figure}

Four healthy users, three male and a single female in their mid 20s participated in the investigation who were all right-handed. Users were provided approximately 5 minutes to practice real and imaginary movements before conducting the recorded sessions prior to the user data collection stage. The visualized movements consisted of moving the right-hand fingers (or arm), left-hand fingers (or arm), kicking, and no movement, corresponding to left yaw turn, right yaw turn, forward, and stationary of a Unitree GO1 robot. The real robot was used throughout data collection, validation, and testing to ensure an accurate control model and consistent brain conditions for the user. Using the real robot through each phase maintained variables like a distracting background, robot vibrations, and lag in wireless video transmission. In the control room, the user faced a monitor displaying the point of view from a camera mounted onboard the robot as displayed in Fig.~\ref{fig_Trio}B and C. Within the Graphical User Interface (GUI), a crosshair was created where an arrow was overlaid corresponding to the MI command for the user. The size of the crosshair was kept small to avoid obstructing the user's view when navigating the robot during real-time MI-BCI control. The user was also asked to focus on the crosshair and avoid looking around the environment via the video feed to avoid eye movement artifacts in the EEG signal. 

\begin{figure*}[!t]
\vspace*{0.2cm}
    \centering
    \includegraphics[width=0.8\linewidth]{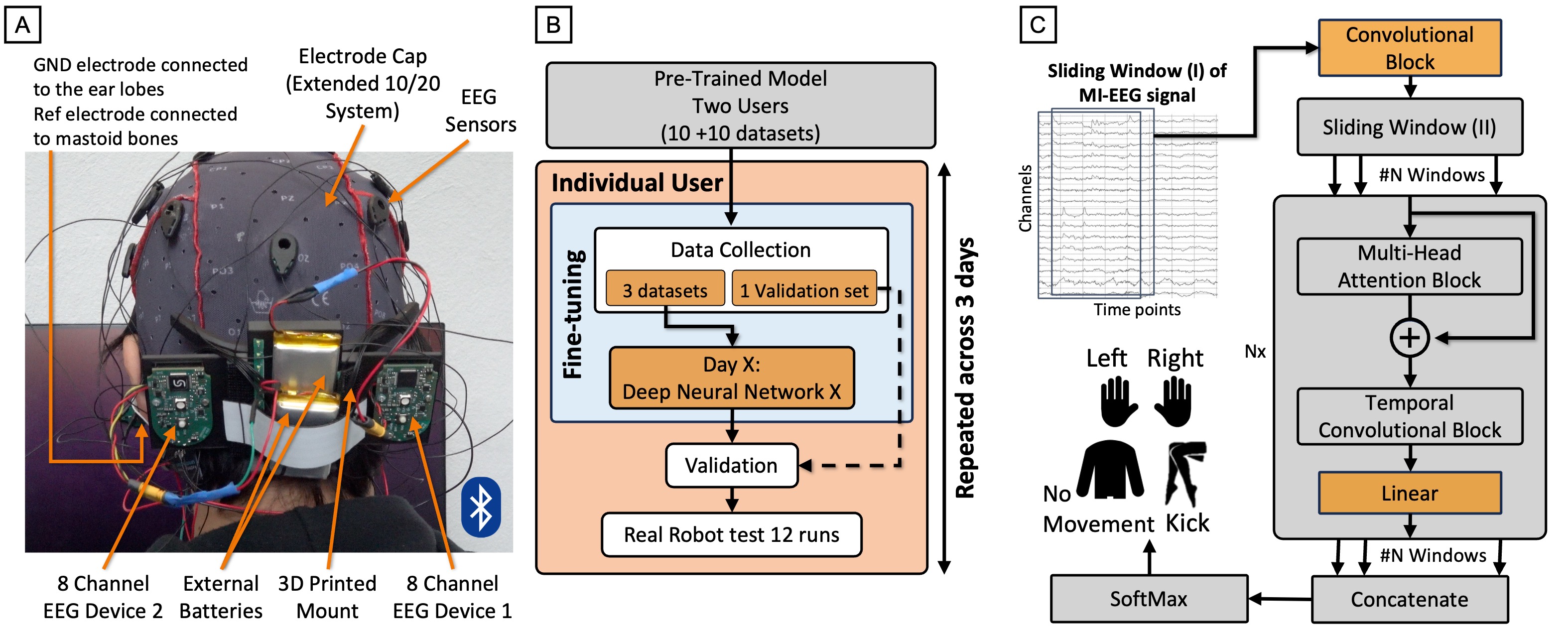}
    \vspace*{0cm}
    \caption{(A) An overview of the hardware setup for the MI-BCI controller. (B) Investigation procedure highlighting each phase of data collection and validation which lead to real robot teleoperation via EEG MI-BCI. (C) An overview of the blocks used to create the DNN where the orange blocks represent the fine-tuned layers and others are frozen. Sliding Window (I) includes the input from ATCNet. Sliding Window (II) is the layer from the original ATCNet model.}
    \label{fig:Procedure_Merged}
    \vspace*{-0.6cm}
\end{figure*}

An overview of the investigation is shown in Fig.~\ref{fig:Procedure_Merged}B to collect the necessary data for the DNN created in this investigation. The pre-trained ATCNet model block included the data collection phase and required a large volume of data to be collected from each user. The procedure listed in order, required 10 datasets (60 runs), 1 validation test (6 runs) and 12 runs controlling the real-world robot for each user. The next step included collecting 3 datasets and 1 validation set to train the fine-tuned model and validate again on the real-world robot for each new day.

In the data collection phase, the randomized labels are sent directly from the GUI to the robot.  For real-robot control testing, EEG signals are processed through the DNN to classify user commands, and are then sent to the real-world robot.

\subsection{Real-time Filtering}
The Direct Current (DC) drift from the raw EEG signals was eliminated by implementing a high-pass Finite Impulse Response (FIR) filter. The principle of the FIR filter involved aggregating a finite series of input signals, each weighted by a designated coefficient, which mirrors the filter's impulse response. This process is mathematically encapsulated by the convolution sum, expressed as:

\begin{equation}
y(n) = \sum_{m=0}^{M-1} x(m) \cdot h(n-m)
\end{equation}
where $y(n)$ denotes the output signal, $x(m)$ refers to the input signal, $h(n-m)$ signifies the impulse response at the time shift of $n-m$, $M$ represents the order of the filter.

The design and optimization of the FIR filter were conducted utilizing the MATLAB tool, \textit{filterDesigner}. For the optimization process, the Least Squares method was selected to refine the filter coefficients, with the order precisely set at $M = 64$. The high pass filter achieved a cutoff frequency of $1.64$ Hz.

\subsection{Input Data to the DNN Decoder}
The DNN decoder received a MI EEG signal $X_i \in \mathbb{R}^{C \times T}$, where $C=16$ represents the EEG electrodes (channels) and $T=1750$ $(250 Hz \times 7 secs)$ the number of time points. The training dataset is a collection of $m$ annotated MI samples $S = \{(X_i, y_i)\}_{i=1}^m$, with $y_i \in \{1, \ldots, n\}$ indicating the class label for trial $X_i$ and $n=4$ denoting the total class count in set $S$. The number of data for each class was balanced to train the DNN. For the pre-training model, a total of $m = 8640$ MI samples was used to train the model. For the fine-tuning model, a total of $m=1296$ MI samples was used to train the model. 

\subsection{Sliding Window on Dataset}
A sliding window was used for real-time decoding of EEG data, updating robot control commands at 1 Hz, as shown in Fig. \ref{fig:Procedure_Merged}C. At any sampling point \( t \), the EEG device samples signals \( \mathbf{f}_t \in \mathbb{R}^C \) from 16 channels. The input vector is constructed as 
\[
\mathbf{X}_{i} = \mathbf{F}_t = [\mathbf{f}_t, \mathbf{f}_{t-1}, \mathbf{f}_{t-2}, \ldots, \mathbf{f}_{t-1749}]^T
\]
which corresponds to EEG signals across 7 seconds. The next data input is \( \mathbf{X}_{i+1} = \mathbf{F}_{t+250}\). The 6-second overlap between consecutive input vectors blends previous and current brain states, enhancing the temporal continuity and capturing more detailed dynamics of neural activity. To train and validate the models, we excluded the first sliding window of data for each MI task, as users might have a delayed response to the MI instruction during the initial second.

\subsection{Deep Neural Network Model}
The DNN model was adapted from a previously established method known as the Attention-based Temporal Convolutional Network (ATCNet) \cite{9852687}. The architecture of the ATCNet model comprises three principal components as shown in Fig.\ref{fig:Procedure_Merged}C: the Convolutional (CV) block, the Attention (AT) block, and the Temporal Convolutional (TC) block. The following parameters of this existing model were adapted to fit our dataset, with the aim to increase the effectiveness of EEG classification.  

In the CV block, the initial layer undertakes a temporal convolution employing $F_1=8$ filters of dimensions $(1, K_C)$, with $K_C=266$ specifying the filter length along the time axis. The output of the CV block, denoted as $z_i \in \mathbb{R}^{T_c \times d}$, presents a temporal sequence of representations. This sequence comprises $T_c=31$ temporal vectors, each of dimension $d = F_2 = F_1 \times D$, where we have empirically set $d$ to 16. The temporal sequence length, $T_c$, is computed as $T_c = T / 8P_2$, where $T$ signifies the time points of the original EEG signal, and $P_2=7$ is a pool length. In the subsequent stage, the convolutional-based sliding window layer segments the temporal sequence $z_i$ into a series of overlapping windows, denoted as $z_i^w \in \mathbb{R}^{T_w \times d}$, where $w = 1, \ldots, n$ represents the index of each window, and $n$ is the total number of such windows. The length of each window, $T_w$, is calculated as $T_w = T_c - n + 1$. For our implementation, we specifically set the sliding window length, $T_w$, to 17, and the total number of windows, $n$, to 15. The AT block has the same parameters as the ACTNet described in \cite{9852687}, except removing the layer norm as it increased performance.

Within the TC block, the Receptive Field Size (RFS) is governed by two parameters: the number of residual blocks, denoted as $L$, and the kernel size, $K_T$. The formula for calculating the RFS is given by 

\begin{equation}
RFS = 1 + 2(K_T - 1)(2L - 1).
\end{equation}

In our application, we set $L$ to 3. Moreover, this TC block employs 12 filters, each with a kernel size of $K_T = 3$ across all convolutional layers. Consequently, this configuration yields an RFS of 21.
We trained the pre-trained model with a large dataset in Day 0 and used 200 epochs. 

We created a pre-trained model using large datasets from two users. For each user and each day, we fine-tuned this pre-trained model with that user's data to create an optimized model. For example, to create the fine-tuned model for User A, we started with the pre-trained model (trained on data from Users A and B) and then optimized it using User A's data for each day.

\subsection{Fine-tuned Model}

Using EEG data, we trained a generalized pre-trained model with 500 epoch to decode motor imagery of 4 classes by combining two users' 10 datasets (E.g. user A and B). The pre-trained model was tested separately on the validation dataset from the same users. Since the model is a convolutional DNN, we fine-tuned the pre-trained model using a small number of datasets compared to the state of the art (3 datasets). This included freezing the central layers of the network for re-training where only the initial CV block and final linear layer have fine-tuned parameters as illustrated in orange in Fig. \ref{fig:Procedure_Merged}C \cite{Peterson}. The fine-tuned model used 15 epochs where a Adam optimization with weight decay of 0.0005 was used with a batch size of 32 and 0.001 learning rate. 

\subsection{Performance Metrics}
The performance of the model is assessed through the metric of accuracy, defined as follows:
\begin{equation}
\mathrm{Accuracy} = \frac{1}{n} \sum_{i=1}^{n} \frac{TP_i}{l_i},
\end{equation}
where $TP_i$ represents the true positives for class $i$, indicating the number of samples correctly identified as belonging to class $i$. The variable $l_i$ denotes the total number of samples within class $i$, and $n$ is the total number of classes. In the case of validation accuracy, all $l_i$ is the same. In the case of robot control accuracy, $i=\textbf{N}$ has more $l_i$.

\section{RESULTS}

\subsection{High Accuracy for Data Collection of Continuous Commands for MI using a Sliding Window} 

\begin{figure}[!b]
    \centering
    \includegraphics[width=1\linewidth]{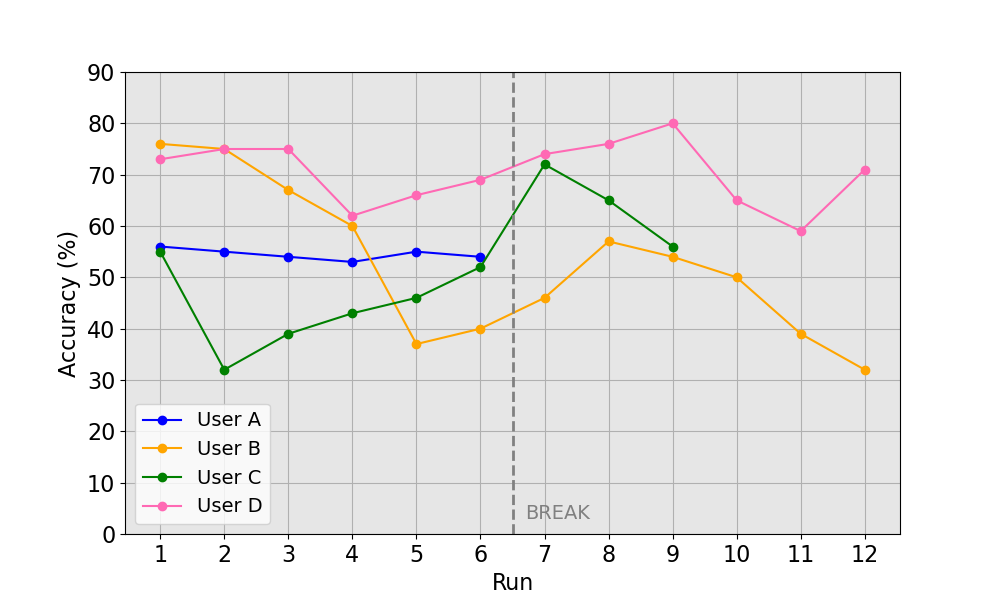}
    \caption{ Accuracy of real-time robot control on Day 0 after long-term data collection, shown as a function of the number of runs. Note that in these robot control datasets, User A-C's concentration decreased compared to the validation datasets due to fatigue, which was mitigated by our fine-tuned approach.}
    \label{DanVsArion_Big_Data_Fig}
\end{figure}

This subsection validates a DNN decoder using a continuous sliding window to decode real-time continuous commands. Table \ref{BCI_Vs_Large_Dataset} shows that the EEG stream of MI was successfully decoded using the DNN model trained by a large number of datasets. The validation accuracy reached 87\%, 67\%, 69\% and 87\% for each user with a total average accuracy of 78\%. The average accuracy of each user included 55\%, 53\%, 51\% and 70\% with a total average accuracy of 57\% for real world MI-BCI control. Changes in decoding accuracy for real-time robot control are represented in Fig.~\ref{DanVsArion_Big_Data_Fig}. The data collection for user A and C was not continued beyond runs 7-12 and 10-12 due to unforeseen circumstances. The drop in decoding accuracy during robot control after long-term data collection (Fig.~\ref{DanVsArion_Big_Data_Fig}, user B) reflects user fatigue. User B's accuracy improved with short breaks but declined again as fatigue set in, highlighting the demanding nature of large dataset collection for DNN training.

\begin{table}[!b]
\caption{Performance of the Large Dataset for Four Users to Control the Robot in Real-time on Day 0}
\label{BCI_Vs_Large_Dataset}

\begin{center}
\scalebox{0.90}{
\begin{tabular}
{|>{\centering\arraybackslash}m{1.5cm}
|>{\centering\arraybackslash}m{0.8cm}
|>{\centering\arraybackslash}m{0.8cm}
|>{\centering\arraybackslash}m{0.8cm}
|>{\centering\arraybackslash}m{0.8cm}
|>{\centering\arraybackslash}m{0.8cm}|}
\hline
 \textbf{Test} & \textbf{User A (\%)} & \textbf{User B (\%)} & \textbf{User C (\%)} & \textbf{User D (\%)} & \textbf{Total Average}\\
\hline
 Validation & 87 & 67 & 69 & 87 & $78 \pm 9.52$ \\
\hline
Robot control & 55 & 53 & 51 & 70 & $57 \pm 7.49$\\
\hline
\end{tabular}
}
\end{center}
\vspace*{-0.5cm}
\end{table}

\begin{table}[!b]
\caption{Performance of the fine-tuned model of Users A-D Across Multiple Days to control the Robot in Real-time}
\label{TL_Table_AD}
\centering
\scalebox{0.90}{
\begin{tabular}
{|>{\centering\arraybackslash}m{1cm}|>{\centering\arraybackslash}m{0.5cm}|>{\centering\arraybackslash}m{0.5cm}|>{\centering\arraybackslash}m{0.5cm}|>{\centering\arraybackslash}m{0.5cm}|>{\centering\arraybackslash}m{0.8cm}|>{\centering\arraybackslash}m{0.8cm}|}
\hline
\textbf{Test} & \textbf{User} & \textbf{Day 1 (\%)} & \textbf{Day 2 (\%)} & \textbf{Day 3 (\%)} & \textbf{Average (\%) (3 days)} & \textbf{Total Average (\%)}\\
\hline

\multirow{4}{*}{\parbox[c][7em][c]{1em}{\rotatebox[origin=c]{90}{Validation}}} & A & 84 & 84 & 80 & $82 \pm 1.89$ & \multirow{4}{*}{\parbox{1cm}{$75 \\ \pm 14.6$}} \\

\cline{2-6} 
& B & 60 & 62 & 58 & $60 \pm 1.63$ &  \\
\cline{2-6} 
& C & 85 & 79 & 63 & $76 \pm 9.28$ & \\
\cline{2-6} 
& D & 82 & 74 & 89 & $82 \pm 6.12$ & \\
\hline

\multirow{4}{*}{\parbox[c][7em][c]{1em}{\rotatebox[origin=c]{90}{Robot Control}}} & A & 74 & 79 & 74 & $76 \pm 2.36$ & \multirow{4}{*}{\parbox{1.5cm}{$62 \\ \pm 16.24$}} \\
\cline{2-6} 
& B & 40 & 47 & 57 & $48 \pm 2.36$ &  \\
\cline{2-6} 
& C & 52 & 50 & 56 & $53 \pm 2.49$ & \\
\cline{2-6} 
& D & 72 & 73 & 74 & $73 \pm 0.82$ & \\
\hline
\end{tabular}
}

\vspace*{-0.1cm}
\end{table}



\begin{figure}[t]
    \centering
    \includegraphics[width=1\linewidth]{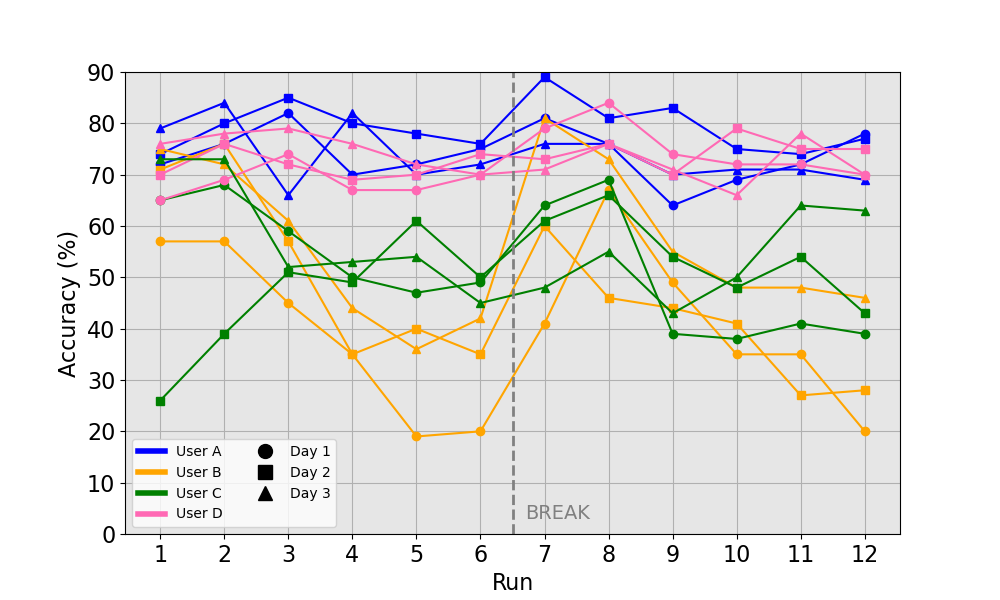}
    \caption{Average accuracy of real-time robot control after short-term data collection on new days, plotted as a function of the number of runs. High accuracy was maintained during robot control due to the DNN optimized for each user and each day.}
    \label{Real_time_Robot_Control_Graph}
    \vspace*{-0.2cm}
\end{figure}

\begin{figure*}[!ht]
    \centering
    \vspace*{0.3cm}
    \includegraphics[width=0.8\linewidth]{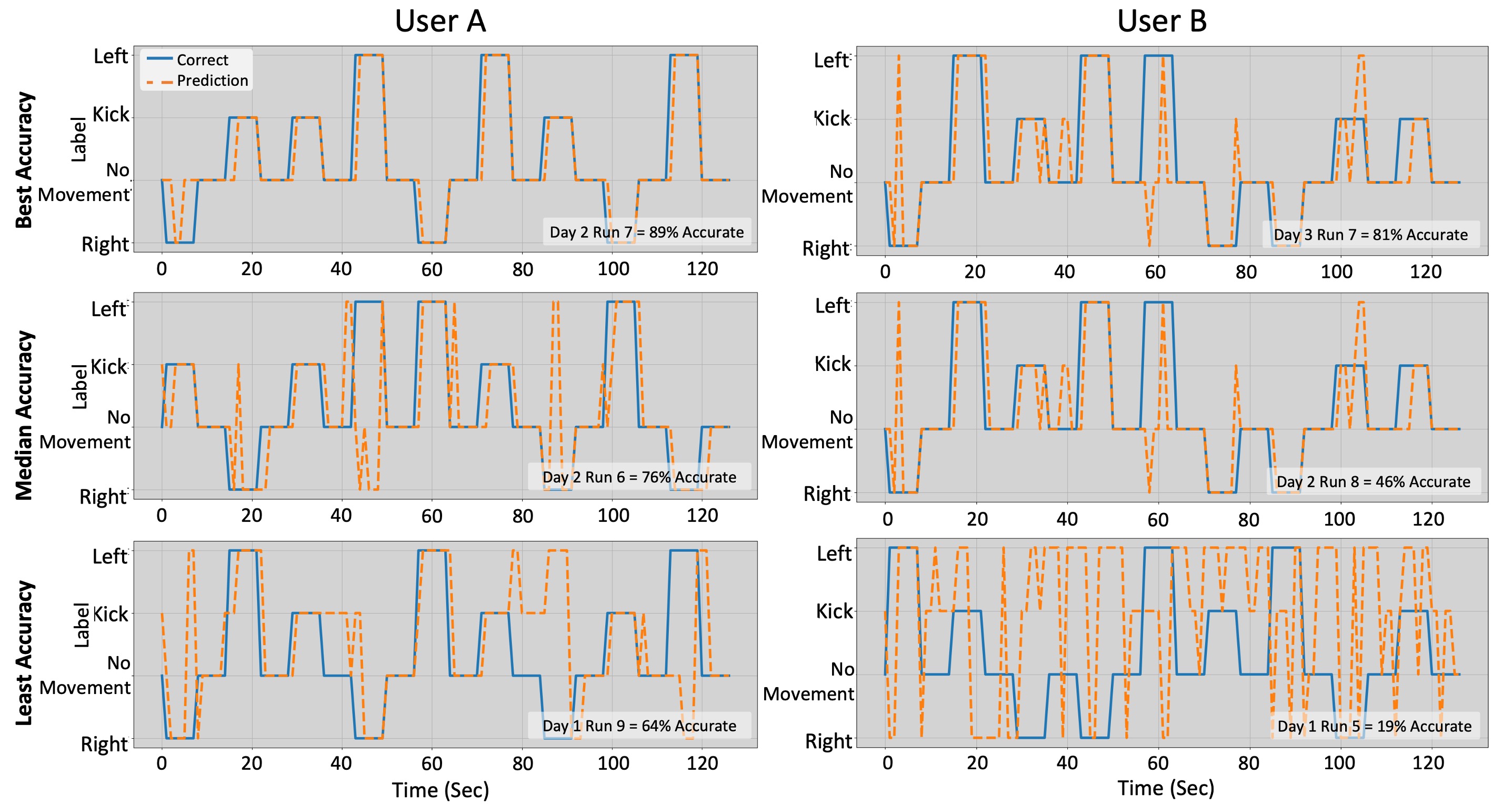}
    \caption{Examples of real-time MI decoding for a robot command for two users. Orange color represents decoded outputs and blue color represents true values.}
    \label{Best_Median_Worst_Graph}
    \vspace*{-0.6cm}
\end{figure*}

\subsection{Optimizing the Decoder with a Small Number of Datasets on New Days}

We examined whether collecting a relatively small number of datasets can lead to high accuracy with the aim to reduce user fatigue when controlling the robot. The pre-trained model from the previous large dataset was fine-tuned to create a new DNN model for each day. The experiment was conducted as follows: Four sessions of data were collected on three different days. Three sessions of data on each day (38 mins total) were used to fine-tune the pre-trained model. The final of the four sessions was used to validate the accuracy of the fine-tuned model for each user. The fine-tuning approach was then repeated and validated across user C and D.

Table \ref{TL_Table_AD}  displays that high decoding accuracy was achieved in the validation test for each user with a total average of 75\% using the fine-tuned model. This ensured high classification accuracy for robot control on each new day whilst reducing the user's mental effort when compared to the large data collection model presented in Section IV.A.

\subsection{Real-Time Robot Control: High Accuracy using Fine-tuned Decoder on New Days}
Based on successful decoding of MI EEG using the fine-tuned DNN model for each new day, we controlled the quadruped using continuous MI in real-time. The GUI randomly displayed commands for user MI-BCI which is then sent to the DNN decoder and robotic platform. 

The examined decoding accuracy was maintained for different days as displayed in Table \ref{TL_Table_AD}. This demonstrates that real-time robot controls were successfully decoded using the fine-tuned method across three different days and validated by two user groups (4 participants), where each achieved a total average accuracy of 62\%. 

Fig. \ref{Real_time_Robot_Control_Graph} displays the results for users A to D for two datasets on three different days. A session is defined as 6 continuous runs lasting 13 minutes total (session 1: runs 1-6, session 2: runs 7-12). A significant rest was taken after the 6th run for the user to refresh. User A demonstrated continuous high accuracy throughout the investigation. However, user B only displayed high accuracy for the first 1-3 and 7-9 runs during the investigation where the accuracy decreased due to user fatigue. High accuracy is regained and attributed to a significant rest indicated in the diagram.

Fig.~\ref{Best_Median_Worst_Graph} shows continuous robot control commands of MI decoding for 6 representative runs for User A and B (a total of 72 runs were conducted for a pair of users). For transparency, we have provided the best, median and least accurate runs to display the MI decoding which resulted in the commands sent to the robot from the BCI device for left, right, no movement and kicking. It can be seen that the maximum accuracy for user A was 89\% and the least accurate was 64\%. User B achieved a maximum of 81\% and the least accurate result of 19\% where user fatigue occurred until a significant rest was taken.


At the end of the investigation, we discovered that User A and D were competent at playing the piano therefore, they found it very intuitive to imagine moving each hand movement when compared to User B and C.

Humans have an average response of 250 \si{\milli\second} for a visual stimuli, in addition to a delay in the camera from the robot. This was taken into consideration for the results where the first sliding window was removed due to human error.

We propose a standardized method for quantitatively evaluating MI commands on real-world mobile robots. While many articles in Table\ref{table_Lit_Review} focus on qualitative assessments of MI-BCI, our approach deploys MI commands on actual robots, enabling quantitative performance assessment under real-world conditions.

\section{CONCLUSION}

In this study, we demonstrated the application of a fine-tuned model for real-time control of a quadruped robot using continuous MI via an affordable BCI device ($\sim$\$3k). The fine-tuned neural network model, using a smaller dataset collected on subsequent days, significantly enhanced real-time robot control. This method maintained high decoding accuracy (75\% validation, 62\% in real robot tests) while reducing user mental load by minimizing data collection requirements by 70\% for each new day. Our findings show that, even with the fine-tuning process, the models achieved robust performance, providing a practical solution to day-to-day variability in EEG signals. Moreover, these results were obtained without formal training beyond a basic understanding of the experimental procedure, underscoring the intuitiveness of MI-BCI. This advancement represents a significant step toward developing user-friendly, efficient BCIs for robot teleoperation. Given the importance of user comfort and system accuracy, the methods presented in this paper will substantially influence the design and usability of future BCIs, particularly for applications requiring continuous and precise control.

\section{Acknowledgement}
The authors would like to express their gratitude to A. Batzorig and J. Arifdjanov for their assistance in validating our approach beyond the letter of acceptance to this conference.

\bibliography{sources}
\bibliographystyle{IEEEtran}

\end{document}